\pdfoutput=1
\documentclass[11pt]{article}

\usepackage[]{ACL2023}

\usepackage{times}
\usepackage{latexsym}
\usepackage{booktabs}  % For improved table formatting
\usepackage{graphicx}  % For the \resizebox command
\usepackage{comment}
\usepackage{enumitem}
\usepackage{multirow}
\usepackage{pgfplots}
\pgfplotsset{compat=1.17}
\usepackage{xcolor}

\usepackage[T1]{fontenc}
\usepackage[utf8]{inputenc}
\usepackage{amssymb}
\usepackage{microtype}

\usepackage{inconsolata}
\usepackage{amsmath}

\title{A Variational Approach for Mitigating Entity Bias in Relation Extraction}
\author{Samuel Mensah$^1$ \quad Elena Kochkina$^1$ \quad Jabez Magomere$^{1,2}$\thanks{\,\,Work done during an internship at JPMorgan AIR} \quad Joy Prakash Sain$^1$ \\ {\bf Simerjot Kaur$^1$ \quad Charese Smiley$^1$} \\ \\
$^1$JP Morgan AI Research \quad $^2$University of Oxford \\
\texttt{\{name\}.\{surname\}@jpmorgan.com} \quad \texttt{jabez.magomere@keble.ox.ac.uk}}

\begin{document}
\maketitle
\begin{abstract}
Mitigating entity bias is a critical challenge in Relation Extraction (RE), where models often rely excessively on entities, resulting in poor generalization. This paper presents a novel approach to address this issue by adapting a Variational Information Bottleneck (VIB) framework. Our method compresses entity-specific information while preserving task-relevant features. It achieves state-of-the-art performance on relation extraction datasets across general, financial, and biomedical domains, in both in-domain (original test sets) and out-of-domain (modified test sets with type-constrained entity replacements) settings. Our approach  offers a robust, interpretable, and theoretically grounded methodology.\footnote{Code available upon request}
\end{abstract}

\section{Introduction}
Relation Extraction (RE) aims to identify and classify semantic relationships between entities in text. For example, to identify an {\it ``investor''} relationship between the entities {\it ``Microsoft''} and {\it ``OpenAI''} in {\it ``Microsoft invests \$10 billion in ChatGPT maker OpenAI''}. %\footnote{Text snippet from \url{https://www.bloomberg.com/news/}} 
By extracting structured relational information from unstructured data, RE serves as a critical enabler for downstream tasks such as knowledge graph construction \cite{distiawan2019neural}, question answering \cite{li2019entity}, and retrieval-augmented generation \cite{lewis2020retrieval}.

While large language models (LLMs), such as LLaMA~\cite{touvron2023llama} and GPT-4~\cite{DBLP:journals/corr/abs-2303-08774}, have been explored for RE tasks~\cite{wei2024chatiezeroshotinformationextraction, li-etal-2023-revisiting-large, zhang-etal-2023-aligning}, fine-tuned pretrained language models (PLMs)  achieve state-of-the-art performance \cite{gutierrez2022thinking,DBLP:conf/emnlp/LiCZPMLS23,zhang2023aligning,wan2023gpt}, particularly in specialized domains like biomedicine~\cite{gutierrez2022thinking} and finance~\cite{DBLP:conf/emnlp/LiCZPMLS23}.
%State-of-the-art RE systems often use pretrained language models (PLMs), such as RoBERTa-Large \cite{liu2019roberta} and LUKE-Large \cite{yamada2020luke}. %, to capture rich contextual information. 

%One of the features ensuring the success of PLMs in RE is entity markers, which identify entities within the input (see @@ and \#\# signs in Figure \ref{fig:motivation-figure1}), and explicitly guide the model's attention to the surrounding context, leading to performance improvements~\cite{DBLP:conf/acl/SoaresFLK19,zhou-chen-2022-improved}.

%The introduction of entity markers \cite{DBLP:conf/acl/SoaresFLK19,zhou-chen-2022-improved}, which identify entities within the input, has led to RE performance by explicitly guiding the model's attention to the surrounding context, where relevant information is typically located. 

\begin{figure}[!ht]
    \centering
    \includegraphics[width=\linewidth]{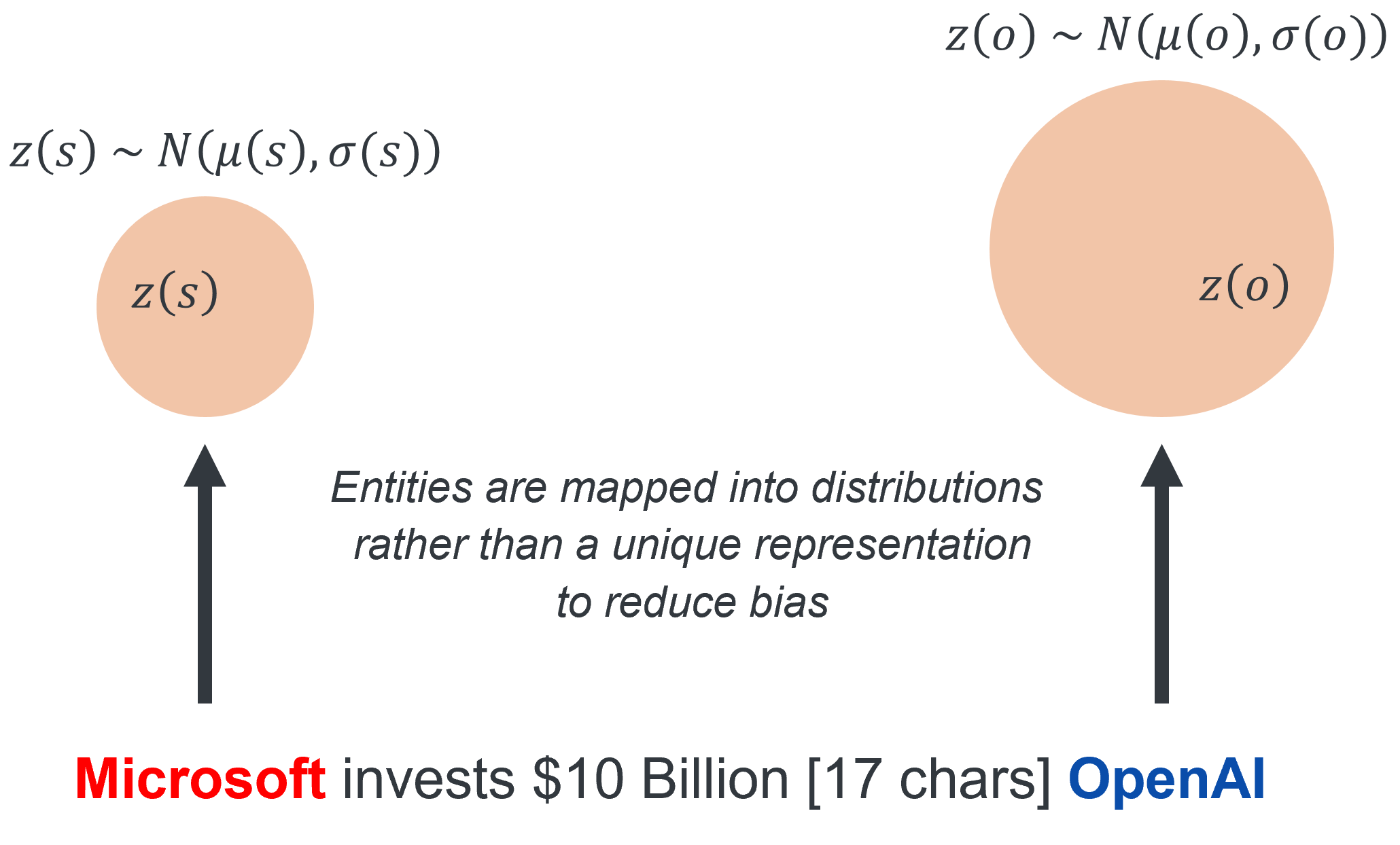}
    \caption{Microsoft, the subject entity $s$ and OpenAI the object entity $o$ are both mapped into stochastic encodings $z(s)$ and $z(o)$ via VIB. The learned variance of the distribution control the variability to reduce bias.}
    \label{fig:intro-fig}
\end{figure}

Despite their success, PLMs often suffer from entity bias \cite{DBLP:conf/emnlp/ZhangZCAM17}, where models overly rely on entity-specific information rather than contextual or relational cues. To mitigate the bias, previous work has explored various solutions, including entity masking~\cite{DBLP:conf/emnlp/ZhangZCAM17,zhang2018graph}, contrastive pre-training~\cite{peng2020learning}, counterfactual analysis~\cite{wang2022should,wang2023extracting} and generation~\cite{modarressi2024consistent}. 

% , thereby neutralizing entity-specific information during inference.

The current state-of-the-art method is a Structured Causal Model (SCM) ~\cite{wang2023causal} that reduces entity bias by constructing a convex hull around an entity's neighbors and using its center to replace the entity embeddings.
%The current state-of-the-art method proposes a Structured Causal Model (SCM)~\cite{wang2023causal} that reduces entity bias by constructing a convex hull around an entity using its neighbors  and replacing the entity embeddings with the center of the convex hull. This helps neutralize entity-specific information.
%and perturbing entity embeddings during training, with the convex hull center used at inference to neutralize entity-specific information. 
%While effective in smoothing the entity representation space and reducing over-reliance on specific entities, SCM lacks the ability to estimate uncertainty, which limits its interpretability. This observation allows us to draw a historical parallel to information bottleneck (IB) principle~\cite{DBLP:journals/corr/physics-0004057}, and in particular the variational information bottleneck (VIB)~\cite{alemi2022deep} %\footnote{Also referred to as variational approach~\cite{alemi2022deep}} that aims to find a compressed representation of the data that preserves relevant information while discarding irrelevant details. 
In contrast to SCM, we draw parallels with variational information bottleneck (VIB)~\cite{alemi2022deep} and propose adapting VIB to map entities to a probabilistic distribution $\mathcal{N}(\mu, \sigma)$, where the variance $\sigma^2$ explicitly quantifies the model’s reliance on entities versus contextual cues. For example, in Fig.~\ref{fig:intro-fig}, the entity {\it Microsoft} is mapped into a tighter distribution compared to {\it OpenAI}. The larger distribution for {\it OpenAI} indicates that the model knows less about it. This helps to debias entities by preventing overconfident assumptions while relying more on the context. Thus this approach not only mitigates entity bias but also enhances interpretability. % and generalization, particularly in out-of-domain scenarios.
%High variance often reflects the model’s effort to rely less on entities and more on context, while low variance indicates greater use of entity information. 
% \jm{This statement is a bit unclear, is high variance related with less confident predictions?}

Our contributions are as follows:
\begin{itemize}[noitemsep, topsep=0pt, leftmargin=*]
\item We propose a novel method for relation extraction, a principled, interpretable, variational framework to reduce entity bias in PLMs, specifically RoBERTa-Large~\cite{liu2019roberta} and LUKE-Large~\cite{yamada2020luke}. 
\item We demonstrate the presence of entity bias in both financial and biomedicine relation extraction domains and compare it with the general domain. 
\item Our method achieves state-of-the-art performance on both general and specialized domains.
\item Our approach’s interpretability is shown through variance analysis, where low variance reflects reliance on entity information, while high variance indicates greater use of context.
%in relations like \texttt{pers:title:title} reflects reliance on entity information, while high variance in \texttt{org:date:formed\_on} indicates greater use of context.
\end{itemize}

\section{Background}
Several works address entity bias in RE through diverse techniques. Entity masking~\cite{DBLP:conf/emnlp/ZhangZCAM17,zhang2018graph}, forces models to focus more on context by replacing entities with generic tokens (e.g., [\texttt{subj-person}]). Entity substitution approaches have been explored to test robustness against entity-based knowledge conflicts in question answering \cite{longpre2021entity} and to mitigate factual bias in document-level RE \cite{modarressi2024consistent}. \citet{peng2020learning} propose to mask entity mentions during pre-training to encourage models to focus on context and type information. \citet{wang2022should} perform counterfactual analysis on a causal graph, to guide models to focus on context without losing entity information. The current SOTA method \cite{wang2023causal} proposes to perturb original entities with neighboring ones to reduce biasing information while preserving context.

Several studies have introduced bias mitigation techniques for black-box large language models (LLMs) that do not require full access to the underlying models (e.g., GPT-4). For instance, \citet{li2024deceptive} demonstrated that LLMs often rely on shortcuts, such as semantic associations or inherent entity biases. \citet{wang2023causal} proposed using an LLM to identify neighboring entities to debias target entities during inference. Similarly, \citet{zhou2023context} showed that employing opinion-based prompts and counterfactual demonstrations can enhance an LLM’s contextual faithfulness. In addition, \citet{zhang2024causal} introduced a causal prompting framework leveraging front-door adjustment to mitigate biases, while \citet{wu2024decot} developed a method for debiasing chain-of-thought reasoning through causal interventions.

Our approach follows the whitebox settings for PLMs. Different from current SOTA method~\cite{wang2023causal}, we debias entities through a probabilitic framework, allowing us to estimate the extent to which we use entity versus contextual information.

\section{Variational Approach}
Our goal is to learn a latent representation $Z$ that preserves the semantic meaning of the input word embeddings $X$, while minimizing the influence of entity information $E$. The variational approach (VIB) \cite{alemi2022deep} provides a principled method to achieve this through the mutual information $I(X;Z|E)$, defined as:\footnote{In this work, $X,Z,E,H$ are random variables, and $x,z,e,h$ are instances of these random variables.} 
% $X$ by minimizing the mutual information $I(X, E)$,
% This can be framed as an information bottleneck problem: maximize the mutual information between X and Z (to preserve semantic information) while minimizing the mutual information between E and Z (to reduce entity bias).
% which is defined as:\footnote{In this work, $X,E,H$ are random variables, and $x,e,h$ are instances of random variables.} 
% \begin{align}
% I(X,E) = &\int dx\,de\, p(x,e) \log \frac{p(x|e)}{p(x)} 
% \end{align}
\vspace{-0.1cm}
\begin{align*}
I(X;Z|E) = &\int dx\, dz\,de\, p(x,z,e) \log \frac{p(z|x,e)}{p(z|e)} \nonumber \\
\leq &\int dx\,dz\, de\, p(x,z,e) \log \frac{p(z|x,e)}{r(z|e)}
\end{align*}
\vspace{-0.1cm}
% \nonumber \\ 
% &\int dh\,de\, p(e)p(h|e) \log \frac{p(h|e)}{r(h)}
where $r(z|e)$ is a variational approximation to $p(z|e)$, inducing an upper bound on $I(X;Z|E)$. We can now interpret the upper bound as a KL divergence \cite{10.1214/aoms/1177729694}, so that the upper bound of $I(X;Z|E)$ becomes the expected KL divergence given by:
% However, directly computing the marginal distribution $p(x)$ is intractable, since it requires integrating over all possible values of $e$. To overcome this, we introduce a variational approximation $r(x)$ to the marginal distribution $p(x)$, leading to the following upper bound n the mutual information:
% \begin{align}
% I(X,E) \leq &\int dx\,de\, p(x,e) \log \frac{p(x|e)}{r(x)} 
% \end{align}
% \footnote{$p(x)=\int de\,p(x|e)p(e)$ requires integrate over all ossible values of $E$ which is intractable}
% where $r(x)$ is a variational approximation to the intractable marginal distribution $p(x)$, typically modelled as a standard normal distribution $\mathcal{N}(0,I)$. $q(x|e)$ is a variational distribution that approximates the true conditional distribution $p(x|e)$. 
% and $q(x|e)$ is a variational distribution that approximates $p(x|e)$
\vspace{-0.1cm}
\begin{align*}
    I(X;Z|E) \leq &\mathbb{E}_{p(x,z,e)}[{\rm KL}(p(z|x,e) || r(z|e))] \nonumber \\
    = &L_{\rm VIB}
\end{align*}
This bound forms the basis of the VIB loss $L_{\rm VIB}$, where the bottleneck is enforced by minimizing the KL divergence, restricting $p(z|x,e)$ to stay close to $r(z|e)$. In practice, $p(z|x, e)$ is modelled as a Gaussian distribution $\mathcal{N}(\mu, \sigma)$, where the mean $\mu$ and  standard deviation $\sigma$ are  parametrized by single-layer perceptrons (SLP)  while $r(z|e)$ is modelled as a standard normal distribution $\mathcal{N}(0,I)$.

% Here, $L_{\rm VIB}$ denotes the VIB loss, which is the KL divergence that measures the difference between the conditional embedding distribution $p(x|e)$ and the variational approximation $r(x)$. Minimizing the KL divergence encourages the embeddings $X$ to contain less entity-specific information of $E$. 
% This effect can be controlled by a multiplier $\beta$, which balances the trade-off between minimizing the classification loss and the regularization imposed by the KL divergence.

% Here, $r(x)$ is a simpler approximating distribution, commonly modeled as a standard normal distribution \mathcal{N}(0,1), which enables the computation of the KL divergence between 

% where $r(h)$ is a variational approximation to the intractable marginal distribution 
% $p(h)$,\footnote{$p(h)=\int de\,p(h|e)p(e)$ requires summing over all possible values of $E$ which is intractable} and this approximation results in an upper bound on the mutual information $I(H,E)$, allowing us to work with a more computationally feasible solution. We can now interpret the upper bound as a KL divergence, so that the upper bound of $I(H,E)$ becomes the expected KL divergence given by:

% \begin{align}
%     I(H,E) \leq \mathbb{E}_{p(e)}[{\rm KL}(p(h|e) || r(h))]
% \end{align}

% Minimizing the KL divergence encourages 
% $p(x|e)$ to retain only the information that is well-represented by $r(h)$, reducing any entity-specific information. This effect can be controlled by a multiplier $\beta$, which balances the trade-off between minimizing the classification loss and the regularization imposed by the KL divergence.
\paragraph{Compressing Entity Representations. }
% To learn compressed representations that limit entity-specific information, we model the distributions $p(x|e)$ and $r(x)$. We simplify $r(x)$ by modeling it as a standard normal distribution, $\mathcal{N}(0, 1)$. This approximation allows for easier computation of the KL divergence between $p(x|e) $ and $r(x)$. We model the distribution $p(x|e)$ as a Gaussian distribution $\mathcal{N}(\mu, \sigma)$ with mean $\mu$ and standard deviation $\sigma$, modelled as linear transformation layers.
% First, let $z$ be the VIB-modified word embeddings, which represent a stochastic, compressed version of the original embedding $x$.
Since the goal is to limit entity-specific information in $Z$ while preserving the semantic meaning in $X$, VIB is applied selectively to entities using a binary entity mask $M$ that identifies the position of entity tokens. To enable efficient and differentiable optimization, we sample $z$ be from $\mathcal{N}(\mu, \sigma)$ using the reparameterization trick~\cite{kingma2013auto}, that is, $z = \mu + \epsilon \cdot \sigma$ and $\epsilon \sim \mathcal{N}(0, 1)$.  $\sigma$ helps control how much information about the input is retained in $z$. Smaller $\sigma$ values lead to tighter, more deterministic representations, while larger $\sigma$ values encourage more stochastic exploration, which is essential for mitigating entity bias and learning better context-based representations.\footnote{To ensure $\sigma>0$, we apply a softplus activation function on the raw ouput $\sigma'$ of the SLP, i.e., $\sigma={\rm softplus}(\sigma')$}

To encourage retaining the context of non-entity tokens and reducing entity-specific details, we selectively blend the original embeddings $x$ with $z$ using $M$ and a blending factor $\beta$. Non-entity tokens $(M = 0)$ retain their original embeddings, while entity tokens $(M = 1)$ are represented as a weighted combination of $x$ and $z$.  
%\resizebox{\linewidth}{!}{
%  \begin{minipage}{0.9\linewidth}
  %\begin{align}
\begin{align*}
x' = x \cdot (1 - M) +
x \cdot M \cdot (1 - \beta) +
z \cdot M \cdot \beta
\end{align*}
%\end{minipage}
%}
This formulation ensures the final embeddings $x'$ reduce entity-specific details while preserving task-relevant features.

\paragraph{Classification and Training Objective.}
Given $x'$, we apply a pretrained PLM encoder to obtain contextualized embeddings $h={\rm PLM}(x')$. We then extract and concatenate the representations of special tags $[h_s]$ and $[h_o]$ which mark the subject and object entities, and feed this joint representation $[h_s;h_o]$ to a fully connected layer and softmax for classification. 
The total loss combines the cross-entropy (CE) loss $L_{\text{CE}}$ for relation classification and the VIB loss $L_{\rm VIB}$.
\vspace{-0.2cm}
\begin{align*}
\mathcal{L} = L_{\text{CE}} + \alpha L_{\text{VIB}}
\end{align*}
where $\alpha$ is an adaptive weight, computed as a ratio between the CE and VIB loss. This ensures a balanced contribution of both loss terms. 

\section{Experiments}
% \paragraph{Experimental Setup} 
We conduct experiments on three large relation extraction datasets: TACRED \cite{zhang2017position} (general domain), REFinD \cite{kaur2023refind} (financial domain) and BioRED~\cite{luo2022biored} (biomedical domain).\footnote{Dataset statistics can be obtained from the original papers.} Evaluation follows previous work~\cite{wang2023causal} using \texttt{entity\_marker\_punctuation}~\cite{zhou-chen-2022-improved} to mark entities, and Micro-F1 as the metric on both in-domain (ID) and out-of-domain (OOD) test sets. Here, in-domain refers to data where entities align with those in the train set, allowing for overlapping entity mentions. Meanwhile, out-of-domain data where entities are replaced to eliminate overlap with the train set. We generate OOD test sets following the approach by \citet{wang2023fragile}, using entities from Wikepedia dumps.\footnote{https://dumps.wikimedia.org/enwiki/latest/enwiki-latest-pages-articles.xml.bz2} We experiment with LUKE-Large~\cite{yamada2020luke} and RoBERTa-Large~\cite{liu2019roberta} as PLM backbones.\footnote{We adapted the source code provided by \citet{wang2023causal} under the Apache-2.0 license. Source code available at \url{https://github.com/luka-group/Causal-View-of-Entity-Bias/} Hyperparameters were tuned for $\beta$ in \{0.1, 0.2 \ldots, 1\}, Learning rates in \{1e-5, 1e-4, 1e-3\}, using Adam as the optimizer. Best hyperparameter $\beta=0.5$ with learning rate $lr=1\text{e-}3$. All experiments were conducted on an AWS g5.24xlarge.}

% \begin{table}[]
%     \centering
%     \begin{tabular}{ccccc} \toprule
%         Dataset&\#Train & \#Dev & \#Test & \#Rel.  \\ \midrule
%          TACRED&68,124&22,631 &15,509 & 42\\ 
%          REFinD&20,070&4,306 &4,300 &22 \\
%          BioRED&& & & \\ \bottomrule
%     \end{tabular}
%     \caption{Dataset Statistics}
%     \label{tab:my_label}
% \end{table}

% We also experiment with RoBERTa-Large \cite{liu2019roberta} and report its results in Appendix~\ref{sec:roberta-large-performance}

%  which was originally used to create the OOD test set for TACRED

% Micro-F1 is used as the evaluation metric, with performance reported as an average over three runs, along with standard deviation to indicate robustness. Evaluation is performed on both in-domain and out-of-domain test sets. 

% The model is based on LUKE-Large~\cite{yamada-etal-2020-luke}, fine-tuned on TACRED and REFinD.  Additionally,  

% Dataset statistics are summarized in Table~\ref{table:dataset_stats}

% % For ID, VIB consistently achieves comparable or superior Micro-F1 across all configurations.  VIB attains an ID Micro-F1 of 70.4\% on TACRED and 75.4\% on REFinD, outperforming SCM by approx. 1.8\% and 0.9\%, respectively. 

\begin{table*}[!ht]
\centering
\footnotesize
\begin{tabular}{lccccccc}
\toprule
& \multicolumn{2}{c}{\textbf{TACRED}} & \multicolumn{2}{c}{\textbf{REFinD}} & \multicolumn{2}{c}{\textbf{BioRED}} \\
\cmidrule(lr){2-3} \cmidrule(lr){4-5} \cmidrule(lr){6-7}
\textbf{Method} & \textsc{\bf ID} & \textsc{\bf OOD} & \textsc{\bf ID} & \textsc{\bf OOD} & \textsc{\bf ID} & \textsc{\bf OOD} \\
\midrule
{\bf LUKE-Large}~\cite{yamada2020luke}            & $71.1_{\pm 0.3}$ & $63.8_{\pm 1.5}$ & $75.0_{\pm 0.2}$ & $73.4_{\pm 0.3}$ & $56.9_{\pm 0.7}$ & $51.8_{\pm 1.2}$ \\ \hline
w/ Ent. Mask~\cite{DBLP:conf/emnlp/ZhangZCAM17}       & $63.6_{\pm 0.1}$ & $61.7_{\pm 1.2}$ & $71.4_{\pm 0.4}$ & $71.4_{\pm 0.9}$ & $53.2_{\pm 0.6}$ & $40.2_{\pm 1.1}$ \\ 
w/ Ent. Substitution~\cite{longpre2021entity}  & $66.6_{\pm 0.3}$ & $60.3_{\pm 0.6}$ & $74.3_{\pm 0.5}$ & $72.9_{\pm 1.2}$ & $56.2_{\pm 0.4}$ & $46.7_{\pm 1.0}$ \\ 
w/ SCM ~\cite{wang2023causal}  & $68.6_{\pm 0.2}$ & $64.8_{\pm 0.4}$ & $74.5_{\pm 0.6}$ & $73.8_{\pm 0.6}$ & $58.3_{\pm 1.7}$ & $53.4_{\pm 1.7}$ \\ \hline
\bf w/ VIB ($\beta=0.5$) & $\mathbf{70.4_{\pm 0.4}}$ & $\mathbf{66.5_{\pm 0.4}}$ & $\mathbf{75.4_{\pm 0.2}}$ & $\mathbf{74.8_{\pm 1.5}}$ & $\mathbf{61.2_{\pm 0.8}}$ & $\mathbf{58.7_{\pm 0.6}}$ \\ 
\midrule
&&&&&& \\
\midrule
{\bf RoBERTa-Large}~\cite{liu2019roberta}            &  $70.8_{\pm 0.1}$    & $61.5_{\pm 0.9}$      &  $75.1_{\pm 0.2}$    &  $72.7_{\pm 0.1}$    & $57.7_{\pm 1.9}$ & $47.9_{\pm 2.3}$  \\ \hline
w/ Entity Mask~\cite{DBLP:conf/emnlp/ZhangZCAM17}            & $62.0_{\pm 0.7}$    &  $60.6_{\pm 0.8}$     &  $70.4_{\pm 1.5}$    &  $71.2_{\pm 1.0}$    & $55.2_{\pm 1.9}$  &  $45.7_{\pm 1.1}$ \\
w/ Entity Substitution~\cite{longpre2021entity}            & $67.1_{\pm 0.3}$    & $61.2_{\pm 1.1}$      & $73.5_{\pm 0.9}$     &  $71.9_{\pm 0.2}$    & $56.9_{\pm 1.1}$  & $46.8_{\pm 3.7}$  \\ 
w/ Structured Causal Model~\cite{wang2023causal}  & $70.5_{\pm 0.6}$    & $\mathbf{67.5_{\pm 0.3}}$       & $74.9_{\pm 1.0}$     & $73.7_{\pm 1.1}$     & $57.3_{\pm 3.3}$  & $\mathbf{52.5_{\pm 3.3}}$ \\ \hline
{\bf w/ VIB} ($\beta=0.5$) &   $\mathbf{70.7_{\pm 0.3}}$  &  $67.2_{\pm 0.3}$   &  $\mathbf{75.4_{\pm 0.1}}$    &  $\mathbf{74.4_{\pm 0.2}}$    & $\mathbf{63.0_{\pm 2.3}}$  & $52.5_{\pm 3.6}$ \\
\bottomrule
% \midrule
\end{tabular}
\caption{{\bf Main Results}: Micro-F1 scores of compared methods with the RoBERTa-Large and LUKE-Large backbones on the TACRED, REFinD, and BioRED datasets, evaluated in both in-domain and out-of-domain settings. Results are averaged over 3 runs, with standard deviations reported.}
\label{tab:results_filled}
\end{table*}

\begin{table*}[!ht]
\centering
\footnotesize
\begin{tabular}{clcl}
\toprule
& \textbf{Var. Bin} & \textbf{Prop.} & \textbf{Dominant Relations (Correct Predictions / Total Gold)} \\
\midrule
\multirow{10}{*}{\rotatebox[origin=c]{90}{LUKE-Large w/ VIB}} & \multicolumn{3}{l}{\textbf{In-Domain}} \\
\cmidrule(lr){2-4}
&0.0-0.1 & 4.6\% & pers:title:title (43/71), org:gpe:headquartered\_in (11/11), org:money:revenue\_of (9/10) \\
&0.1-0.2 & 85.8\% & pers:title:title (503/600), org:gpe:operations\_in (419/536), pers:org:employee\_of (329/352) \\
&0.2-0.3 & 9.6\% & org:date:formed\_on (73/78), org:gpe:operations\_in (55/60), org:org:subsidiary\_of (4/6) \\
&0.3-0.4 & 0.1\% & org:date:formed\_on (3/3) \\
\cmidrule(lr){2-4}
&\multicolumn{3}{l}{\textbf{Out-of-Domain}} \\
\cmidrule(lr){2-4}
&0.0-0.1 & 13.2\% & pers:title:title (59/107), pers:org:employee\_of (49/89), org:gpe:operations\_in (19/30) \\
&0.1-0.2 & 82.8\% & pers:title:title (463/564), org:gpe:operations\_in (433/550), pers:org:employee\_of (259/283) \\
&0.2-0.3 & 3.8\% & org:date:formed\_on (66/68), org:gpe:operations\_in (22/25), pers:org:employee\_of (2/2) \\
&0.3-0.4 & 0.2\% & org:date:formed\_on (8/8) \\
\midrule
\midrule
% \end{tabular}
% % \caption{Variance analysis on REFinD in-domain and out-of-domain test sets, grouped by mean variance. Results are based on the VIB model with a LUKE-Large backbone.}
% \caption{Distribution of samples across variance bins in REFinD in-domain and out-of-domain test sets, with dominant relations identified for each bin. Results are based on the VIB model with the LUKE-Large backbone.}
% \label{tab:uncertainty_analysis}
% \end{table*}

% \begin{table*}[htbp]
% \centering
% \footnotesize
% \begin{tabular}{lcl}
% \toprule
% \textbf{Variance Range} & \textbf{Pct} & \textbf{Dominant Relations (Correct Predictions / Total Gold)} \\
\multirow{14}{*}{\rotatebox[origin=c]{90}{RoBERTa-Large w/ VIB}} & \multicolumn{3}{l}{\textbf{In-Domain}} \\
\cmidrule(lr){2-4}

% \multirow{14}{*}{{\centering\rotatebox[origin=c]{90}{RoBERTa-Large w/VIB}}}&&& \\
% \midrule
% &\multicolumn{3}{l}{\textbf{In-Domain}} \\
% \midrule
&0.0-0.1 & 13.9\% & pers:title:title (476/479), pers:org:employee\_of (31/40), org:gpe:operations\_in (12/14) \\
&0.1-0.2 & 43.4\% & pers:org:employee\_of (287/297), org:gpe:operations\_in (280/332), pers:title:title (75/181) \\
&0.2-0.3 & 35.0\% & org:gpe:operations\_in (157/222), pers:org:employee\_of (34/37), org:org:agreement\_with (24/74) \\
&0.3-0.4 & 6.4\% & org:date:formed\_on (57/60), org:gpe:operations\_in (25/34), org:org:subsidiary\_of (5/10) \\
&0.4-0.5 & 1.0\% & org:date:formed\_on (15/15), org:gpe:headquartered\_in (1/1), org:gpe:operations\_in (1/2) \\
&0.5-0.6 & 0.1\% & org:date:formed\_on (1/1), org:gpe:operations\_in (1/1), org:org:subsidiary\_of (1/1) \\
\cmidrule(lr){2-4}
&\multicolumn{3}{l}{\textbf{Out-of-Domain}} \\
\cmidrule(lr){2-4}
&0.0-0.1 & 15.0\% & pers:title:title (424/442), pers:org:employee\_of (49/75), org:gpe:operations\_in (33/40) \\
&0.1-0.2 & 56.3\% & org:gpe:operations\_in (299/397), pers:org:employee\_of (254/277), pers:title:title (102/221) \\
&0.2-0.3 & 24.8\% & org:gpe:operations\_in (114/151), pers:org:employee\_of (19/21), org:date:formed\_on (16/20) \\
&0.3-0.4 & 3.2\% & org:date:formed\_on (53/55), org:gpe:operations\_in (12/17), org:money:revenue\_of (1/2) \\
&0.4-0.5 & 0.7\% & org:date:formed\_on (12/12), no\_relation (0/0), org:org:shares\_of (0/4) \\
&0.5-0.6 & 0.0\% & org:date:formed\_on (1/1) \\
\bottomrule
\end{tabular}
\caption{Variance analysis of REFinD ID and OOD test sets, categorized by variance bins (Var. Bin). The table highlights the proportion of samples (Prop.) within each bin and identifies dominant relations based on correct predictions versus total gold labels. Results are presented for both LUKE- and RoBERTa-Large 
 w/VIB models.}
% \caption{Variance analysis on REFinD in-domain and out-of-domain test sets, grouped by mean variance (i.e., Var. Bin).}
\label{tab:uncertainty_analysis}
\end{table*}

\section{Main Results}
%\subsection{Performance}
The results in Table~\ref{tab:results_filled} highlight the performance of LUKE-Large and RoBERTa-Large backbone models. We find that traditional methods like Entity Masking~\cite{DBLP:conf/emnlp/ZhangZCAM17} and Entity Substitution~\cite{longpre2021entity} show underperformance, highlighting the importance of retaining some information about the original entity. Both SCM and VIB retain some information about the original entity, leading to their stronger performance compared to early methods. 

For LUKE-Large in ID settings, VIB achieves Micro-F1 scores: 70.4\% on TACRED, 75.4\% on REFinD and 61.2\% on BioRED, outperforming SCM by about 1.8\%, 0.9\% and 2.9\%, respectively. Under entity-replaced conditions (OOD), VIB consistently shows competitive or better performance compared to SCM. Specifically, VIB achieves Micro F1 scores: 66.5\% on TACRED, 74.8\% on REFinD, and 58.7\% on BioRED, outperforming SCM by about 1.7\%, 1\% and 5.3\%, respectively. For the RoBERTa-Large backbone, SCM and VIB achieve comparable performance, with SCM slightly outperforming VIB in OOD TACRED (67.5\% vs. 67.2\%) and OOD BioRED (52.5\% vs. 52.5\%). VIB has an edge in ID and OOD REFinD, and ID for both TACRED and BioRED. 

Comparing the results of VIB across the backbones, LUKE’s knowledge-based entity representations appear to amplify VIB’s ability to effectively balance the utilization of entities and context. This highlights VIB’s strength in leveraging entity-rich backbones for improved generalization, especially in domain-specific datasets like REFinD.

\subsection{Variance Analysis}
% \jm{From a reader's perspective, it is quite unclear how variance \(\sigma^2\) shows up, given that the VIB loss is explained initially using \(\sigma\), maybe we can add a line clarifying that we obtain the variance for each test instance based on model output estimations during a forward pass? I assume the variance is based on the learned weights from the SLP in VIB loss}
 During inference, $\sigma$ is predicted by the learned SLP that parameterizes the distribution $\mathcal{N}(\mu, \sigma)$, with variance computed as $\sigma^2$.
Variance $\sigma^2$ reflects the model's reliance on entities versus context, where low variance indicates stronger reliance on entities and high variance reflects greater use of contextual cues. We analyze Micro-F1 performance and data distribution across in-domain and out-of-domain settings on REFinD to understand this balance in the financial domain.

\begin{figure}
    \centering
    \includegraphics[width=\linewidth]{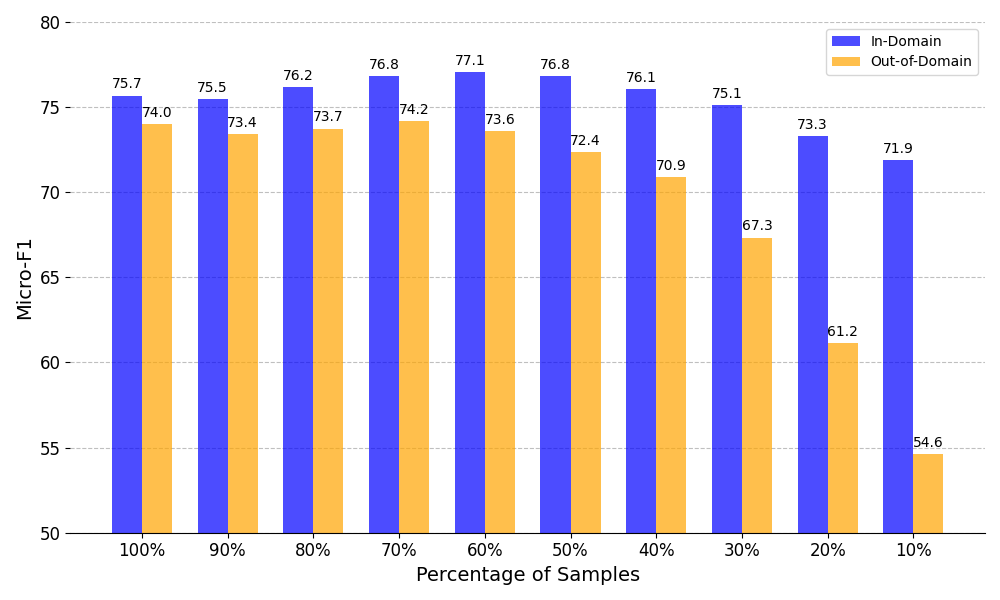}
    \caption{Micro-F1 scores across sample subsets (sorted by variance) for ID and OOD on REFinD.}
    \label{fig:micro-f1-disbn}
\end{figure}

\textbf{Micro-F1 Distribution Across Variance}
Figure~\ref{fig:micro-f1-disbn} illustrates VIB's Micro-F1 scores for in-domain and out-of-domain datasets. Samples are grouped by ascending mean variance, where subsets with lower percentages (e.g., 10\%) correspond to the highest variance. For in-domain data, the scores remain stable across subsets (75.7\% to 71.9\%), indicating that VIB effectively mitigates entity bias while leveraging both entity and contextual information. In out-of-domain data, however, the sharper decline in Micro-F1 scores (74.0\% to 54.6\%) indicates that while VIB reduces over-reliance on entities by mapping entities into distributions of high variances, contextual signals may not always provide strong predictive cues. This underscores the importance of robust context-entity interaction for generalization.

\textbf{Data Distribution Across Variance} 
In Table~\ref{tab:uncertainty_analysis}, we group instances into bins based on mean variance - average variance across entity tokens. Each bin reports the proportion of samples, and the dominant relations, highlighting the most accurately predicted relations.
The results highlight VIB’s ability to balance entity and contextual information while adapting to varying data distributions in in-domain and out-of-domain settings. For in-domain data, most samples (85.8\%) fall into the 0.1–0.2 variance bin, dominated by relations like \texttt{pers:title:title} and \texttt{org:gpe:operations\_in}, with smaller proportions in lower (0.0–0.1, 4.6\%) and moderate (0.2–0.3, 9.6\%) variance bins. This concentrated distribution explains the stability of Micro-F1 scores observed in the bar graph, as removing high-variance samples has minimal impact on performance. In contrast, out-of-domain data shifts more samples into the lowest variance bin (0.0–0.1, 13.2\%), reflecting stronger reliance on entities; however, entity replacements disrupt predictive utility, leading to lower performance in the bar graph. Additionally, sparsely populated high-variance bins (e.g., 0.2–0.3, 3.8\%) correspond to sharp performance drops (e.g., 30\%–10\%), highlighting challenges with relying on context alone and the need for stronger contextual adaptability in out-of-domain scenarios (see examples in~\ref{sec:appendix_mask}).

\section{Conclusions}
We proposed a novel robust, interpretable, and theoretically grounded method for mitigating entity bias in relation extraction. We evaluated this approach on general and domain-specific datasets, TACRED, REFinD and BioRED, and showed that it achieves state-of-the-art results on each.

\section*{Limitations}
In this study, we focus on the application of PLMs, acknowledging that our work does not easily extend to LLMs, which have become increasingly significant in recent advancements. Future research will aim to expand our VIB method to encompass generative models such as T5 and LLMs, potentially uncovering new insights and applications. 

% Additionally, our analysis is confined to a single domain-specific dataset, which may restrict the generalizability of our results to other fields. Despite our belief in the potential for broader applicability, further validation across diverse datasets, such as those in biomedical relation extraction (BioRE) and other domains, is necessary. 

Furthermore, our research is conducted solely in the English language, which may limit its relevance to non-English contexts. Language-specific challenges and nuances could influence the performance of PLMs, and future studies should consider incorporating multiple languages to enhance the generalizability and impact of our findings. 

%\section*{Ethical Considerations}
%We use publicly available datasets that dont have PI information. 
%We use classification models that 

\section*{Acknowledgments}
We thank Arturo Oncevay, Toyin
Aguda, Keshav Ramani and Xiaomo Liu for insightful discussions.

\paragraph*{Disclaimer} This paper was prepared for informational purposes by the Artificial Intelligence Research group of JPMorgan Chase \& Co. and its affiliates ("JP Morgan'') and is not a product of the Research Department of JP Morgan. JP Morgan makes no representation and warranty whatsoever and disclaims all liability, for the completeness, accuracy or reliability of the information contained herein. This document is not intended as investment research or investment advice, or a recommendation, offer or solicitation for the purchase or sale of any security, financial instrument, financial product or service, or to be used in any way for evaluating the merits of participating in any transaction, and shall not constitute a solicitation under any jurisdiction or to any person, if such solicitation under such jurisdiction or to such person would be unlawful.

\bibliography{anthology,custom}

\begin{thebibliography}{35}
\expandafter\ifx\csname natexlab\endcsname\relax\def\natexlab#1{#1}\fi

\bibitem[{Alemi et~al.(2022)Alemi, Fischer, Dillon, and Murphy}]{alemi2022deep}
Alexander~A Alemi, Ian Fischer, Joshua~V Dillon, and Kevin Murphy. 2022.
\newblock Deep variational information bottleneck.
\newblock In \emph{International Conference on Learning Representations}.

\bibitem[{Distiawan et~al.(2019)Distiawan, Weikum, Qi, and Zhang}]{distiawan2019neural}
Bayu Distiawan, Gerhard Weikum, Jianzhong Qi, and Rui Zhang. 2019.
\newblock Neural relation extraction for knowledge base enrichment.
\newblock In \emph{Proceedings of the 57th Annual Meeting of the Association for Computational Linguistics}, pages 229--240.

\bibitem[{Guti{\'e}rrez et~al.(2022)Guti{\'e}rrez, McNeal, Washington, Chen, Li, Sun, and Su}]{gutierrez2022thinking}
Bernal~Jim{\'e}nez Guti{\'e}rrez, Nikolas McNeal, Clayton Washington, You Chen, Lang Li, Huan Sun, and Yu~Su. 2022.
\newblock Thinking about gpt-3 in-context learning for biomedical ie? think again.
\newblock In \emph{Findings of the Association for Computational Linguistics: EMNLP 2022}, pages 4497--4512.

\bibitem[{Kaur et~al.(2023)Kaur, Smiley, Gupta, Sain, Wang, Siddagangappa, Aguda, and Shah}]{kaur2023refind}
Simerjot Kaur, Charese Smiley, Akshat Gupta, Joy Sain, Dongsheng Wang, Suchetha Siddagangappa, Toyin Aguda, and Sameena Shah. 2023.
\newblock Refind: Relation extraction financial dataset.
\newblock In \emph{Proceedings of the 46th International ACM SIGIR Conference on Research and Development in Information Retrieval}, pages 3054--3063.

\bibitem[{Kingma(2013)}]{kingma2013auto}
Diederik~P Kingma. 2013.
\newblock Auto-encoding variational bayes.
\newblock \emph{arXiv preprint arXiv:1312.6114}.

\bibitem[{Kullback and Leibler(1951)}]{10.1214/aoms/1177729694}
Solomon Kullback and Richard~A. Leibler. 1951.
\newblock \href {https://doi.org/10.1214/aoms/1177729694} {{On Information and Sufficiency}}.
\newblock \emph{The Annals of Mathematical Statistics}, 22(1):79 -- 86.

\bibitem[{Lewis et~al.(2020)Lewis, Perez, Piktus, Petroni, Karpukhin, Goyal, K{\"u}ttler, Lewis, Yih, Rockt{\"a}schel et~al.}]{lewis2020retrieval}
Patrick Lewis, Ethan Perez, Aleksandra Piktus, Fabio Petroni, Vladimir Karpukhin, Naman Goyal, Heinrich K{\"u}ttler, Mike Lewis, Wen-tau Yih, Tim Rockt{\"a}schel, et~al. 2020.
\newblock Retrieval-augmented generation for knowledge-intensive nlp tasks.
\newblock \emph{Advances in Neural Information Processing Systems}, 33:9459--9474.

\bibitem[{Li et~al.(2024)Li, Zhou, Wang, Fu, Roth, and Chen}]{li2024deceptive}
Bangzheng Li, Ben Zhou, Fei Wang, Xingyu Fu, Dan Roth, and Muhao Chen. 2024.
\newblock Deceptive semantic shortcuts on reasoning chains: How far can models go without hallucination?
\newblock In \emph{Proceedings of the 2024 Conference of the North American Chapter of the Association for Computational Linguistics: Human Language Technologies (Volume 1: Long Papers)}, pages 7668--7681.

\bibitem[{Li et~al.(2023{\natexlab{a}})Li, Wang, and Ke}]{li-etal-2023-revisiting-large}
Guozheng Li, Peng Wang, and Wenjun Ke. 2023{\natexlab{a}}.
\newblock \href {https://aclanthology.org/2023.findings-emnlp.459} {Revisiting large language models as zero-shot relation extractors}.
\newblock In \emph{Findings of the Association for Computational Linguistics: EMNLP 2023}, pages 6877--6892, Singapore. Association for Computational Linguistics.

\bibitem[{Li et~al.(2023{\natexlab{b}})Li, Chan, Zhu, Pei, Ma, Liu, and Shah}]{DBLP:conf/emnlp/LiCZPMLS23}
Xianzhi Li, Samuel Chan, Xiaodan Zhu, Yulong Pei, Zhiqiang Ma, Xiaomo Liu, and Sameena Shah. 2023{\natexlab{b}}.
\newblock \href {https://doi.org/10.18653/V1/2023.EMNLP-INDUSTRY.39} {Are chatgpt and {GPT-4} general-purpose solvers for financial text analytics? {A} study on several typical tasks}.
\newblock In \emph{Proceedings of the 2023 Conference on Empirical Methods in Natural Language Processing: {EMNLP} 2023 - Industry Track, Singapore, December 6-10, 2023}, pages 408--422. Association for Computational Linguistics.

\bibitem[{Li et~al.(2019)Li, Yin, Sun, Li, Yuan, Chai, Zhou, and Li}]{li2019entity}
Xiaoya Li, Fan Yin, Zijun Sun, Xiayu Li, Arianna Yuan, Duo Chai, Mingxin Zhou, and Jiwei Li. 2019.
\newblock Entity-relation extraction as multi-turn question answering.
\newblock In \emph{Proceedings of the 57th Annual Meeting of the Association for Computational Linguistics}, pages 1340--1350.

\bibitem[{Liu et~al.(2019)Liu, Ott, Goyal, Du, Joshi, Chen, Levy, Lewis, Zettlemoyer, and Stoyanov}]{liu2019roberta}
Yinhan Liu, Myle Ott, Naman Goyal, Jingfei Du, Mandar Joshi, Danqi Chen, Omer Levy, Mike Lewis, Luke Zettlemoyer, and Veselin Stoyanov. 2019.
\newblock Roberta: A robustly optimized bert pretraining approach.
\newblock \emph{arXiv preprint arXiv:1907.11692}.

\bibitem[{Longpre et~al.(2021)Longpre, Perisetla, Chen, Ramesh, DuBois, and Singh}]{longpre2021entity}
Shayne Longpre, Kartik Perisetla, Anthony Chen, Nikhil Ramesh, Chris DuBois, and Sameer Singh. 2021.
\newblock Entity-based knowledge conflicts in question answering.
\newblock In \emph{Proceedings of the 2021 Conference on Empirical Methods in Natural Language Processing}, pages 7052--7063.

\bibitem[{Luo et~al.(2022)Luo, Lai, Wei, Arighi, and Lu}]{luo2022biored}
Ling Luo, Po-Ting Lai, Chih-Hsuan Wei, Cecilia~N Arighi, and Zhiyong Lu. 2022.
\newblock Biored: a rich biomedical relation extraction dataset.
\newblock \emph{Briefings in Bioinformatics}, 23(5):bbac282.

\bibitem[{Modarressi et~al.(2024)Modarressi, K{\"o}ksal, and Sch{\"u}tze}]{modarressi2024consistent}
Ali Modarressi, Abdullatif K{\"o}ksal, and Hinrich Sch{\"u}tze. 2024.
\newblock Consistent document-level relation extraction via counterfactuals.
\newblock In \emph{Findings of the Association for Computational Linguistics: EMNLP 2024}, pages 11501--11507.

\bibitem[{OpenAI(2023)}]{DBLP:journals/corr/abs-2303-08774}
OpenAI. 2023.
\newblock \href {https://doi.org/10.48550/ARXIV.2303.08774} {{GPT-4} technical report}.
\newblock \emph{CoRR}, abs/2303.08774.

\bibitem[{Peng et~al.(2020)Peng, Gao, Han, Lin, Li, Liu, Sun, and Zhou}]{peng2020learning}
Hao Peng, Tianyu Gao, Xu~Han, Yankai Lin, Peng Li, Zhiyuan Liu, Maosong Sun, and Jie Zhou. 2020.
\newblock Learning from context or names? an empirical study on neural relation extraction.
\newblock In \emph{Proceedings of the 2020 Conference on Empirical Methods in Natural Language Processing (EMNLP)}, pages 3661--3672.

\bibitem[{Sun et~al.(2019)Sun, Zhang, Mensah, Mao, and Liu}]{sun2019aspect}
Kai Sun, Richong Zhang, Samuel Mensah, Yongyi Mao, and Xudong Liu. 2019.
\newblock Aspect-level sentiment analysis via convolution over dependency tree.
\newblock In \emph{Proceedings of the 2019 conference on empirical methods in natural language processing and the 9th international joint conference on natural language processing (EMNLP-IJCNLP)}, pages 5679--5688.

\bibitem[{Touvron et~al.(2023)Touvron, Lavril, Izacard, Martinet, Lachaux, Lacroix, Rozi{\`e}re, Goyal, Hambro, Azhar et~al.}]{touvron2023llama}
Hugo Touvron, Thibaut Lavril, Gautier Izacard, Xavier Martinet, Marie-Anne Lachaux, Timoth{\'e}e Lacroix, Baptiste Rozi{\`e}re, Naman Goyal, Eric Hambro, Faisal Azhar, et~al. 2023.
\newblock Llama: Open and efficient foundation language models.
\newblock \emph{arXiv preprint arXiv:2302.13971}.

\bibitem[{Wan et~al.(2023)Wan, Cheng, Mao, Liu, Song, Li, and Kurohashi}]{wan2023gpt}
Zhen Wan, Fei Cheng, Zhuoyuan Mao, Qianying Liu, Haiyue Song, Jiwei Li, and Sadao Kurohashi. 2023.
\newblock Gpt-re: In-context learning for relation extraction using large language models.
\newblock In \emph{Proceedings of the 2023 Conference on Empirical Methods in Natural Language Processing}, pages 3534--3547.

\bibitem[{Wang et~al.(2023{\natexlab{a}})Wang, Mo, Wang, Zhou, and Chen}]{wang2023causal}
Fei Wang, Wenjie Mo, Yiwei Wang, Wenxuan Zhou, and Muhao Chen. 2023{\natexlab{a}}.
\newblock A causal view of entity bias in (large) language models.
\newblock In \emph{Findings of the Association for Computational Linguistics: EMNLP 2023}, pages 15173--15184.

\bibitem[{Wang et~al.(2023{\natexlab{b}})Wang, Zhang, Deng, Gardner, Roth, and Chen}]{wang2023extracting}
Haoyu Wang, Hongming Zhang, Yuqian Deng, Jacob Gardner, Dan Roth, and Muhao Chen. 2023{\natexlab{b}}.
\newblock Extracting or guessing? improving faithfulness of event temporal relation extraction.
\newblock In \emph{Proceedings of the 17th Conference of the European Chapter of the Association for Computational Linguistics}, pages 541--553.

\bibitem[{Wang et~al.(2022)Wang, Chen, Zhou, Cai, Liang, Liu, Yang, Liu, and Hooi}]{wang2022should}
Yiwei Wang, Muhao Chen, Wenxuan Zhou, Yujun Cai, Yuxuan Liang, Dayiheng Liu, Baosong Yang, Juncheng Liu, and Bryan Hooi. 2022.
\newblock Should we rely on entity mentions for relation extraction? debiasing relation extraction with counterfactual analysis.
\newblock In \emph{Proceedings of the 2022 Conference of the North American Chapter of the Association for Computational Linguistics: Human Language Technologies}, pages 3071--3081.

\bibitem[{Wang et~al.(2023{\natexlab{c}})Wang, Hooi, Wang, Cai, Liang, Zhou, Tang, Duan, and Chen}]{wang2023fragile}
Yiwei Wang, Bryan Hooi, Fei Wang, Yujun Cai, Yuxuan Liang, Wenxuan Zhou, Jing Tang, Manjuan Duan, and Muhao Chen. 2023{\natexlab{c}}.
\newblock How fragile is relation extraction under entity replacements?
\newblock In \emph{Proceedings of the 27th Conference on Computational Natural Language Learning (CoNLL)}, pages 414--423.

\bibitem[{Wei et~al.(2024)Wei, Cui, Cheng, Wang, Zhang, Huang, Xie, Xu, Chen, Zhang, Jiang, and Han}]{wei2024chatiezeroshotinformationextraction}
Xiang Wei, Xingyu Cui, Ning Cheng, Xiaobin Wang, Xin Zhang, Shen Huang, Pengjun Xie, Jinan Xu, Yufeng Chen, Meishan Zhang, Yong Jiang, and Wenjuan Han. 2024.
\newblock \href {http://arxiv.org/abs/2302.10205} {Chatie: Zero-shot information extraction via chatting with chatgpt}.

\bibitem[{Wu et~al.(2024)Wu, Yu, Chen, Wang, Rossi, Kim, Rao, and McAuley}]{wu2024decot}
Junda Wu, Tong Yu, Xiang Chen, Haoliang Wang, Ryan Rossi, Sungchul Kim, Anup Rao, and Julian McAuley. 2024.
\newblock Decot: Debiasing chain-of-thought for knowledge-intensive tasks in large language models via causal intervention.
\newblock In \emph{Proceedings of the 62nd Annual Meeting of the Association for Computational Linguistics (Volume 1: Long Papers)}, pages 14073--14087.

\bibitem[{Yamada et~al.(2020)Yamada, Asai, Shindo, Takeda, and Matsumoto}]{yamada2020luke}
Ikuya Yamada, Akari Asai, Hiroyuki Shindo, Hideaki Takeda, and Yuji Matsumoto. 2020.
\newblock Luke: Deep contextualized entity representations with entity-aware self-attention.
\newblock In \emph{Proceedings of the 2020 Conference on Empirical Methods in Natural Language Processing (EMNLP)}, pages 6442--6454.

\bibitem[{Zhang et~al.(2024)Zhang, Zhang, Wu, Zhou, and He}]{zhang2024causal}
Congzhi Zhang, Linhai Zhang, Jialong Wu, Deyu Zhou, and Yulan He. 2024.
\newblock Causal prompting: Debiasing large language model prompting based on front-door adjustment.
\newblock \emph{arXiv preprint arXiv:2403.02738}.

\bibitem[{Zhang et~al.(2023{\natexlab{a}})Zhang, Guti{\'e}rrez, and Su}]{zhang2023aligning}
Kai Zhang, Bernal~Jim{\'e}nez Guti{\'e}rrez, and Yu~Su. 2023{\natexlab{a}}.
\newblock Aligning instruction tasks unlocks large language models as zero-shot relation extractors.
\newblock In \emph{Findings of the Association for Computational Linguistics: ACL 2023}, pages 794--812.

\bibitem[{Zhang et~al.(2023{\natexlab{b}})Zhang, Jimenez~Gutierrez, and Su}]{zhang-etal-2023-aligning}
Kai Zhang, Bernal Jimenez~Gutierrez, and Yu~Su. 2023{\natexlab{b}}.
\newblock \href {https://doi.org/10.18653/v1/2023.findings-acl.50} {Aligning instruction tasks unlocks large language models as zero-shot relation extractors}.
\newblock In \emph{Findings of the Association for Computational Linguistics: ACL 2023}, pages 794--812, Toronto, Canada. Association for Computational Linguistics.

\bibitem[{Zhang et~al.(2018)Zhang, Qi, and Manning}]{zhang2018graph}
Yuhao Zhang, Peng Qi, and Christopher~D Manning. 2018.
\newblock Graph convolution over pruned dependency trees improves relation extraction.
\newblock In \emph{Proceedings of the 2018 Conference on Empirical Methods in Natural Language Processing}, pages 2205--2215.

\bibitem[{Zhang et~al.(2017{\natexlab{a}})Zhang, Zhong, Chen, Angeli, and Manning}]{DBLP:conf/emnlp/ZhangZCAM17}
Yuhao Zhang, Victor Zhong, Danqi Chen, Gabor Angeli, and Christopher~D. Manning. 2017{\natexlab{a}}.
\newblock \href {https://doi.org/10.18653/V1/D17-1004} {Position-aware attention and supervised data improve slot filling}.
\newblock In \emph{Proceedings of the 2017 Conference on Empirical Methods in Natural Language Processing, {EMNLP} 2017, Copenhagen, Denmark, September 9-11, 2017}, pages 35--45. Association for Computational Linguistics.

\bibitem[{Zhang et~al.(2017{\natexlab{b}})Zhang, Zhong, Chen, Angeli, and Manning}]{zhang2017position}
Yuhao Zhang, Victor Zhong, Danqi Chen, Gabor Angeli, and Christopher~D Manning. 2017{\natexlab{b}}.
\newblock Position-aware attention and supervised data improve slot filling.
\newblock In \emph{Conference on empirical methods in natural language processing}.

\bibitem[{Zhou and Chen(2022)}]{zhou-chen-2022-improved}
Wenxuan Zhou and Muhao Chen. 2022.
\newblock \href {https://doi.org/10.18653/v1/2022.aacl-short.21} {An improved baseline for sentence-level relation extraction}.
\newblock In \emph{Proceedings of the 2nd Conference of the Asia-Pacific Chapter of the Association for Computational Linguistics and the 12th International Joint Conference on Natural Language Processing (Volume 2: Short Papers)}, pages 161--168, Online only. Association for Computational Linguistics.

\bibitem[{Zhou et~al.(2023)Zhou, Zhang, Poon, and Chen}]{zhou2023context}
Wenxuan Zhou, Sheng Zhang, Hoifung Poon, and Muhao Chen. 2023.
\newblock Context-faithful prompting for large language models.
\newblock In \emph{Findings of the Association for Computational Linguistics: EMNLP 2023}, pages 14544--14556.

\end{thebibliography}
\bibliographystyle{acl_natbib}

\appendix
\section{Appendix}
\label{sec:appendix}

% \subsection{Datasets}
% We evaluate our approach on three widely relation extraction datasets: TACRED (general domain), REFinD (financial domain), BioRED (biomedicine domain). Given the 512-token context limitation of LUKE-Large or RoBERTa-Large, we exclude instances where the subject or object entity is positioned beyond the 512 limit, following the same setup as previous work~\cite{wang2023causal}. The resulting dataset statistics is presented in Table \ref{tab:dataset-stats}.

% \begin{table}[htbp]
%     \centering
%     \footnotesize
%     \begin{tabular}{ccccc}
%     \toprule
%          \bf Dataset & \bf \#Train &\bf  \#Test-ID &\bf  \#Test-OOD &\bf  \#Rel.  \\ \midrule
%          TACRED& 68,124  & 15,509 & 12,419 & 42\\
%          REFinD& 20,045  & 4,294 & 4,294 & 22\\
%          BioRED& 3,246  & 836 & 831 & 8\\
%          \bottomrule
%     \end{tabular}
%     \caption{Dataset Statistics. }
%     \label{tab:dataset-stats}
% \end{table}

\subsection{Performance Across Different Relations}
\label{sec:appendix:perfbreakdown}
\textbf{REFinD} Table~\ref{tab:performance_comparison-luke-large-refind-breakdown} shows a detailed comparison of SCM and VIB performance across multiple relations in REFinD, evaluated in both ID and OOD settings. VIB outperforms SCM in many cases, particularly for relations like \texttt{org:org:agreement\_with} and \texttt{pers:org:member\_of}, where contextual cues are critical, demonstrating its ability to effectively balance entity and context. For high-frequency relations such as \texttt{no\_relation} and \texttt{org:gpe:operations\_in}, VIB also shows slight but consistent improvements. However, SCM often matches or outperforms VIB on relations like \texttt{org:money:loss\_of} and \texttt{org:gpe:headquartered\_in}. While both methods achieve comparable overall performance, VIB provides the added advantage of quantifying the reliance on entity versus context information, making it more insightful for understanding model behavior.

\textbf{TACRED} Table \ref{tab:luke_performance} compares SCM and VIB performance on TACRED in both ID and OOD settings, across various relation types. While SCM excels in high-frequency relations like \texttt{no\_relation} and \texttt{per:title}, VIB consistently outperforms SCM in challenging relations such as \texttt{per:employee\_of} and \texttt{org:top\_members/employees}, particularly under entity replacement in OOD. VIB's strength lies in leveraging contextual information effectively when entity reliability diminishes. For rare relations like \texttt{per:city\_of\_death} and \texttt{org:dissolved}, VIB often surpasses SCM, though both methods struggle with extremely sparse relations like \texttt{org:shareholders}. Overall, VIB demonstrates strong generalization under entity replacement, making it a robust approach for mitigating entity bias while maintaining competitive performance across diverse relations.

\begin{table}[!ht]
\centering
\footnotesize
\begin{tabular}{lcccc}
\toprule
& \multicolumn{2}{c}{\textbf{REFinD-ID}} & \multicolumn{2}{c}{\textbf{REFinD-OOD}} \\
\cmidrule(lr){2-3} \cmidrule(lr){4-5}
\textbf{Relation} & \textbf{SCM} & \textbf{VIB} & \textbf{SCM} & \textbf{VIB} \\
\midrule
no\_relation & 85.01 & \bf 86.91 & 85.01 & \bf 86.91 \\
pers:title:title & 77.79 & 77.79 & 77.79 & 77.79 \\
org:gpe:operations & 76.20 & \bf 78.35 & 76.20 & \bf 78.35 \\
pers:org:employee & \bf 93.05 & 82.89 & \bf 93.05 & 82.89 \\
org:org:agrmnt & 26.95 & \bf 35.46 & 26.95 & \bf 35.46 \\
org:date:formed & 86.32 & \bf 87.37 & 86.32 & \bf 87.37 \\
pers:org:member & 9.47 & \bf 15.79 & 9.47 & \bf 15.79 \\
org:org:subsidiary & 38.55 & \bf 49.40 & 38.55 & \bf 49.40 \\
org:org:shares & \bf 27.87 & 6.56 & \bf 27.87 & 6.56 \\
org:money:revenue & 74.47 & \bf 82.98 & 74.47 & \bf 82.98 \\
org:money:loss & \bf 96.77 & 90.32 & \bf 96.77 & 90.32 \\
org:gpe:headqtr & 79.31 & 79.31 & 79.31 & 79.31 \\
org:date:acquired & \bf 54.17 & 37.50 & \bf 54.17 & 37.50 \\
pers:org:founder & \bf 40.00 & 30.00 & \bf 40.00 & 30.00 \\
org:gpe:formed & 23.53 & \bf 64.71 & 23.53 & \bf 64.71 \\
pers:univ:employee & 58.33 & \bf 66.67 & 58.33 & \bf 66.67 \\
org:org:acquired & \bf 18.18 & 0.00 & \bf 18.18 & 0.00 \\
pers:gov:member & \bf 12.50 & 0.00 & \bf 12.50 & 0.00 \\
pers:univ:attended & 85.71 & 85.71 & 85.71 & 85.71 \\
org:money:profit & 80.00 & 80.00 & 80.00 & 80.00 \\
pers:univ:member\ & \bf 60.00 & 40.00 & \bf 60.00 & 40.00 \\
org:money:cost & \bf 75.00 & 0.00 & \bf 75.00 & 0.00 \\
\bottomrule
\end{tabular}
\caption{{\bf LUKE-Large} Performance of SCM and VIB models on various relations within the {\bf REFinD dataset}, evaluated in both in-domain and out-of-domain settings. Relations are ordered by their frequency in the dataset, with the most frequent at the top (i.e., no\_relation). Bolded values indicate the best performance for a relation in either ID or OOD settings.}
\label{tab:performance_comparison-luke-large-refind-breakdown}
\end{table}

\begin{table}[!ht]
\centering
\small
\begin{tabular}{lcccc}
\toprule
& \multicolumn{2}{c}{\textbf{TACRED-ID}} & \multicolumn{2}{c}{\textbf{TACRED-OOD}} \\
\cmidrule(lr){2-3} \cmidrule(lr){4-5}
\textbf{Relation} & \textbf{SCM} & \textbf{VIB} & \textbf{SCM} & \textbf{VIB} \\
\midrule
no\_relation & \bf 93.71 & 92.61 & \bf 93.60 & 92.26 \\
per:title & \bf 94.20 & 89.60 & \bf 93.90 & 92.20 \\
org:top\_memb/empl & 80.64 & \bf 82.66 & 74.28 & \bf 76.30 \\
per:employee & 53.41 & \bf 71.59 & 38.64 & \bf 55.30 \\
org:alt\_names & 91.08 & \bf 91.55 & \bf 80.75 & 78.87 \\
per:age & \bf 96.00 & 95.50 & \bf 97.63 & 97.63 \\
per:cities\_res & 48.68 & \bf 57.14 & 40.31 & \bf 54.26 \\
per:countries\_res & 5.41 & \bf 43.92 & 5.77 & \bf 48.08 \\
per:origin & \bf 65.91 & 48.48 & \bf 65.79 & 54.39 \\
org:country\_of\_hq & 37.04 & \bf 58.33 & 30.84 & \bf 38.32 \\
per:charges & 88.35 & \bf 93.20 & \bf 91.25 & 91.25 \\
per:parents & \bf 79.55 & 79.55 & \bf 79.52 & 78.31 \\
org:city\_hq & \bf 73.17 & 67.07 & \bf 67.07 & 65.85 \\
per:state\_res & 49.38 & \bf 59.26 & 45.61 & \bf 49.12 \\
org:founded\_by & \bf 86.76 & 85.29 & \bf 82.35 & 80.88 \\
per:spouse & 50.00 & \bf 78.79 & 46.77 & \bf 75.81 \\
org:parents & 35.48 & \bf 43.55 & 20.97 & \bf 24.19 \\
per:other\_fam & \bf 51.67 & 51.67 & \bf 58.82 & 56.86 \\
per:siblings & 67.27 & \bf 76.36 & 72.55 & \bf 76.47 \\
per:date\_death & 18.52 & \bf 44.44 & 23.08 & \bf 58.97 \\
per:cause\_death & 40.38 & \bf 48.08 & 42.50 & \bf 50.00 \\
org:state\_hq & \bf 76.47 & 74.51 & \bf 74.51 & 72.55 \\
per:religion & \bf 42.55 & 40.43 & \bf 51.61 & 48.39 \\
org:subsidiaries & 40.91 & \bf 45.45 & \bf 36.36 & 34.09 \\
org:founded & \bf 83.78 & 83.78 & \bf 86.11 & 80.56 \\
per:children & 40.54 & \bf 43.24 & 40.63 & \bf 50.00 \\
org:members & 0.00 & 0.00 & 0.00 & \bf 3.23 \\
per:sch\_attended & 60.00 & \bf 83.33 & 46.67 & \bf 70.00 \\
per:city\_death & 0.00 & \bf 39.29 & 0.00 & \bf 52.17 \\
org:website & \bf 84.62 & 84.62 & 41.67 & \bf 79.17 \\
org:num\_empl/memb & \bf 68.42 & 57.89 & \bf 70.59 & 54.90 \\
org:member & 0.00 & 0.00 & 0.00 & 0.00 \\
per:state\_death & 0.00 & \bf 42.86 & 0.00 & \bf 42.42 \\
org:shareholders & 0.00 & 0.00 & 0.00 & 0.00 \\
per:alt\_names & \bf 27.27 & 18.18 & \bf 21.21 & 6.06 \\
org:pol/relig\_affil & \bf 40.00 & 40.00 & \bf 48.15 & 44.44 \\
per:date\_birth & \bf 77.78 & 77.78 & \bf 75.00 & 75.00 \\
per:country\_death & 0.00 & 0.00 & 0.00 & 0.00 \\
per:state\_birth & 25.00 & \bf 50.00 & 27.78 & \bf 50.00 \\
per:country\_birth & 0.00 & \bf 20.00 & 0.00 & \bf 26.67 \\
per:city\_birth & 20.00 & \bf 40.00 & 20.00 & \bf 40.00 \\
org:dissolved & 0.00 & \bf 50.00 & 0.00 & \bf 50.00 \\
\bottomrule
\end{tabular}
\caption{{\bf Luke-Large} Performance of models SCM and VIB on various relations within {\bf TACRED dataset}, evaluated in both in-domain and out-of-domain settings. Relations are ordered by their frequency in the dataset, with the most frequent at the top (i.e., no\_relation). Bolded values indicate the best performance for a relation in either ID or OOD settings.}
\label{tab:luke_performance}
\end{table}

\begin{figure*}[!ht]
    \centering
    \includegraphics[width=\linewidth]{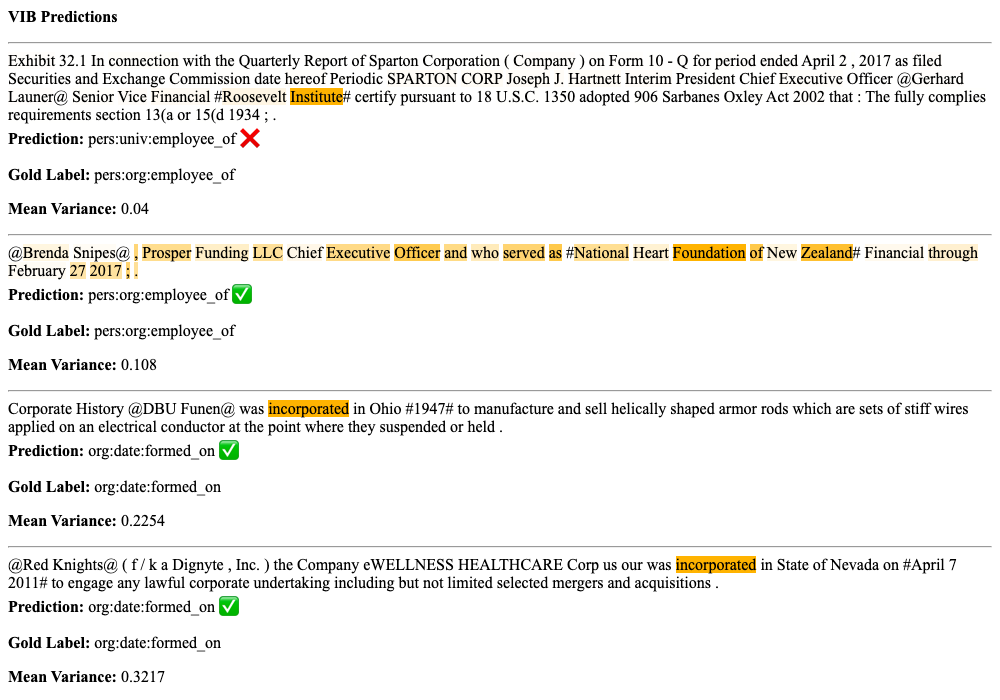}
    \caption{Visualization of attention and prediction results for VIB. Subject and object entities are marked with @ and \# respectively. Low mean variance indicates strong reliance on entity tokens, while high mean variance reflects a shift toward contextual cues. Highlighted tokens show entity-focused attention, visualized using the mask method~\cite{sun2019aspect}}
    \label{fig:motivation-figure1}
\end{figure*}

\subsection{Mask Experiment Vrs Variance Analysis}
\label{sec:appendix_mask}
The mask experiment, as proposed by \cite{sun2019aspect}, evaluates token-level relevance by measuring the contribution of individual tokens to the final relation representation, effectively visualizing the model’s attention patterns. Complementary to this, variance analysis provides a quantitative measure of reliance on entity or contextual information, where low variance indicates strong reliance on entity tokens and high variance reflects greater dependence on contextual cues. As demonstrated in Figure~\ref{fig:motivation-figure1}, entities in relations like \texttt{pers:org:employee\_of} exhibit low mean variance (e.g., 0.108), aligning with the mask experiment’s focus on entity tokens such as ``@Brenda Snipes@'' and ``\#Prosper Funding LLC\#''. Conversely, for relations like \texttt{org:date:formed\_on}, higher mean variance (e.g., 0.2254 and 0.3217) suggests greater reliance on context, consistent with the mask experiment, where contextual words like ``incorporated'' contribute prominently. This alignment between variance analysis and the mask experiment highlights the model’s ability to balance entity and contextual cues, reinforcing  interpretability.

\end{document}